%% file: main.tex
\documentclass{article}

\usepackage{spconf,amsmath,graphicx,amssymb}
\usepackage{hyphenat}
\usepackage{verbatim}
\usepackage{subfig,float}
\usepackage{multirow}
\usepackage{xr-hyper}
\usepackage{hyperref}

\usepackage[]{algorithm2e}

\title{Heteroscedastic Calibration of Uncertainty Estimators in Deep Learning}
\name{Bindya Venkatesh$^{\dagger}$ and Jayaraman J. Thiagarajan$^{\ddagger}$\thanks{This work was performed under the auspices of the U.S. Department of Energy by Lawrence Livermore National Laboratory under Contract DE-AC52-07NA27344.}}
\address{$^{\dagger}$Arizona State University, $^{\ddagger}$ Lawrence Livermore National Labs}
\begin{document}
%
\maketitle
\begin{abstract}
The role of uncertainty quantification (UQ) in deep learning has become crucial with growing use of predictive models in high-risk applications. Though a large class of methods exists for measuring deep uncertainties, in practice, the resulting estimates are found to be poorly calibrated, thus making it challenging to translate them into actionable insights. A common workaround is to utilize a separate recalibration step, which adjusts the estimates to compensate for the miscalibration. Instead, we propose to repurpose the heteroscedastic regression objective as a surrogate for calibration and enable any existing uncertainty estimator to be inherently calibrated. In addition to eliminating the need for recalibration, this also regularizes the training process. Using regression experiments, we demonstrate the effectiveness of the proposed heteroscedastic calibration with two popular uncertainty estimators.
\end{abstract}

\section{Introduction}
\label{sec:intro}
\input{intro.tex}

\section{Background}
\label{sec:background}
\input{background.tex}

\section{Proposed Approach}
\label{sec:approach}
\input{approach.tex}

\section{Empirical Results}
\label{sec:experiments}
\input{results.tex}
\section{Conclusions}
\label{sec:conclusions}
\input{conclusions.tex}

\bibliographystyle{IEEEbib}
\bibliography{refs}

\end{document}

%% file: intro.tex
The use of deep learning models in critical applications such as healthcare and autonomous driving has made it imperative to characterize model reliability and to support interpretability. Since a variety of factors pertinent to data sampling, measurement errors and model approximation contribute to the stochasticity in data-driven methods, uncertainty quantification (UQ) has been found to be essential for studying model behavior~\cite{montavon2018methods,doshi2017towards,thiagarajan2019understanding}. While UQ methods have been widely adopted in several statistical learning frameworks~\cite{smith2013uncertainty}, there has been a recent surge in interest to generalize UQ methods to deep learning~\cite{gal2016dropout,lakshminarayanan2017simple,tagasovska2018frequentist,gal2017concrete, kendall2017uncertainties}. Due to the lack of application-specific priors on the uncertainties, we often utilize metrics such as \textit{calibration}, which quantifies the likelihood of containing the true target in the estimated prediction intervals, to evaluate the quality of uncertainty estimates.

A natural strategy to produce calibrated predictors is to directly optimize for prediction intervals that satisfy the calibration objective~\cite{thiagarajan2019building}. For example, in the heteroscedastic regression~\cite{gal2016uncertainty} approach, the variance estimates are obtained using the Gaussian likelihood objective, under a heteroscedastic prior assumption. However, by not explicitly constructing the intervals based on epistemic (model variability) or aleatoric (inherent stochasticity) uncertainties, it is not straightforward to interpret the variances from a heteroscedastic model, even when they are well calibrated. On the other hand, approaches designed to capture specific sources of uncertainties, e.g. Monte Carlo dropout for epistemic~\cite{gal2016dropout} or conditional quantile based aleatoric uncertainties~\cite{tagasovska2018frequentist}, are found to be poorly calibrated in practice~\cite{kuleshov2018accurate}. Hence, a typical workaround is to employ a separate recalibration step that adjusts the estimates from a trained model to achieve calibration~\cite{guo2017calibration,platt1999probabilistic, niculescu2005predicting,kuleshov2018accurate}. However, it was showed recently in~\cite{levi2019evaluating} that even uninformative (random) interval estimates can be effectively recalibrated, thus rendering the estimates meaningless for subsequent analysis. 

Instead, we propose to repurpose the heteroscedastic regression objective as a surrogate for calibration, and enable any existing uncertainty estimator to produce inherently calibrated intervals. In other words, with this single-shot calibration approach, the uncertainty estimates are used in lieu of the heteroscedastic variances to compute the Gaussian likelihood. By performing calibration automatically in the training process based on an explicit uncertainty estimator, our approach does not suffer the limitations of recalibration methods~\cite{levi2019evaluating} and can be associated to specific error sources unlike classical heteroscedastic networks. Surprisingly, our approach is able to achieve significantly improved calibration with both an epistemic (MC dropout) and an aleatoric (quantile-based) uncertainty estimator, though they are known to be produce miscalibrated intervals in practice~\cite{thiagarajan2019building}. More importantly, this implicit calibration objective regularizes the training process and produces highly accurate mean estimators.

%% file: background.tex
\RestyleAlgo{boxruled}
\begin{algorithm}[t]
	
	\KwIn{Labeled data $\{(\mathrm{x}_i, \mathrm{y}_i)\}_{i=1}^N$, epochs $T$, Monte-Carlo iterations $M$, dropout rate $p$.}
	\KwOut{Trained model that produces predictions with intervals}
	\textbf{Initialization}:Randomly initialize model parameters\;
	
	\For{$T$ epochs}{
	    \For{$M$ iterations}{
	         Perform forward pass with dropout $\hat{\mathrm{y}}_i^j = \mathcal{F}(\mathrm{x}_i; \text{dropout} = p) \quad \forall i$ \;
	    }
	    Estimate mean $\mathrm{\mu}_i = \frac{1}{M} \sum_{j=1}^M \hat{\mathrm{y}}_i^j \quad \forall i$\;
	    Estimate variance $\mathrm{\sigma}_i^2 = \text{Var}[\hat{\mathrm{y}}_i^j] \quad \forall i$ \;
		Compute heteroscedastic regression objective using Eq. (\ref{eq:hnn}) \;
		Update $\Theta^* = \displaystyle \arg \min_{\Theta} \mathrm{L}_{\text{HNN}} $ \;
		}
	\caption{Dropout-HC: MC Dropout (epistemic) uncertainty estimator with heteroscedastic calibration}\label{algo-d}
\end{algorithm}

\RestyleAlgo{boxruled}
\begin{algorithm}[t]
	\KwIn{Labeled data $\{(\mathrm{x}_i, \mathrm{y}_i)\}_{i=1}^N$, epochs $T$, upper quantile $\tau^u$, lower quantile $\tau^l$.}
	\KwOut{Trained model that produces predictions with intervals} 
	\textbf{Initialization}:Randomly initialize model parameters\;
	
	\For{$T$ epochs}{
	    Compute mean prediction $\mathrm{\mu}_i = \mathcal{F}(\mathrm{x}_i) \quad \forall i$\;
	    Compute $\hat{\mathrm{y}}_i^u = \mathcal{F}(\mathrm{x}_i|\tau = \tau^u) \quad \forall i$ \;
	    Compute $\hat{\mathrm{y}}_i^l = \mathcal{F}(\mathrm{x}_i|\tau = \tau^l) \quad \forall i$ \;
	    Estimate stddev $\mathrm{\sigma}_i = \frac{\hat{\mathrm{y}}_i^u - \hat{\mathrm{y}}_i^l}{2} \quad \forall i$ \;
		Compute heteroscedastic regression objective using Eq. (\ref{eq:hnn}) \;
		Update $\Theta^* = \displaystyle \arg \min_{\Theta} \lambda_H\mathrm{L}_{\text{HNN}} + \lambda_u \mathrm{L}_{\tau^u} + \lambda_l \mathrm{L}_{\tau^l}$ \;
		}
	\caption{Quantile-HC: Quantile-based aleatoric uncertainty estimator with heteroscedastic calibration}\label{algo-q}
\end{algorithm}

For an input $\mathrm{x} \subset \mathcal{X} \in \mathbb{R}^d $, a model $\mathcal{F}$ with parameters $\Theta$, maps the input space $\mathcal{X}$ to the output space $\mathcal{Y}$, i.e., $ \mathcal{F}: \mathrm{x} \mapsto \mathrm{y}$, where $\mathrm{y} \subset \mathcal{Y} \in \mathbb{R}$. Our goal is to enable models to produce inherently calibrated intervals for each prediction, in lieu of point estimates.

\paragraph*{Heteroscedastic Regression:} The proposed approach relies on repurposing the heteroscedastic regression objective as a surrogate for calibration, with any given uncertainty estimator. In heteroscedastic regression, it is assumed that the prediction for each sample $\mathrm{x}_i$ is modeled as a Gaussian $\mathcal{N}(\mathrm{\mu}_i, \mathrm{\sigma}_i^2)$, wherein both means and variances are estimated by the model $\mathcal{F}$. The model parameters are then optimized with the Gaussian likelihood objective~\cite{gal2016uncertainty,nix1994estimating}:
\begin{equation} \label{eq:hnn}
    \mathrm{L}_{\text{HNN}}(\mathrm{y}_i,\mathrm{\mu}_i,\mathrm{\sigma}_i) = 
\frac{(\mathrm{y}_i - \mathrm{\mu}_i)^2}{2\mathrm{\sigma}_i^2} + \frac{1}{2} \log(\mathrm{\sigma}_i^2).
\end{equation}We refer to this model as a heteroscedastic neural network (HNN). Note that, unlike classical UQ methods, this black-box method directly produces prediction intervals, thus making it challenging to interpret what error sources it measures.

\paragraph*{Calibration:} This is a widely adopted metric for evaluating intervals in predictive models~\cite{gneiting2007probabilistic}. In classification tasks, the estimated uncertainties in terms of class probabilities are well calibrated if the class probability assigned is consistent with the prediction accuracy~\cite{guo2017calibration}, i.e., the model should be less confident about a wrong prediction. In regression tasks, the prediction intervals are considered to be well calibrated if the probability of the true target falling in the interval matches the true empirical probability~\cite{kuleshov2018accurate}. Formally, given the estimates $(\mathrm{\mu}, \mathrm{\sigma})$ for a sample $\mathrm{x}$, the prediction interval is represented as
\begin{equation}\label{eq:interval}
[\hat{\mathrm{y}}^l, \hat{\mathrm{y}}^u] = [\mathrm{\mu} - z_{(1-\alpha)/2}{\mathrm{\sigma}},\; \mathrm{\mu} + z_{(1-\alpha)/2}{\mathrm{\sigma}}].
\end{equation}Here, $\alpha$ denotes the desired level of calibration and $z$ indicates the $z-$score (for instance, $z_{(1-\alpha)/2} = 1.96,$ for $95\%$ calibration). 
We quantify how well the intervals are calibrated using the \textit{calibration error} (CE) metric~\cite{levasseur2017uncertainties}:
\begin{equation}\label{eq:calib}
\text{CE} = \displaystyle \sum_{\alpha \in A} \left|\alpha - \frac{1}{N}{ \sum_{i=1}^{N}\mathbb{I}\left[\hat{\mathrm{y}}_{i}^l \leq \mathrm{y}_i \leq \hat{\mathrm{y}}_{i}^u\right]} \right|.
\end{equation}Here, $A$ is the set of desired calibration levels (set to [0.1, 0.3, 0.5, 0.7, 0.9, 0.99] in our experiments), $N$ is the total number of samples, and $\mathbb{I}$ denotes the indicator function.

\begin{figure*}[t]
	\centering
	\subfloat[S][Boston - Dropout Rate vs CE]{\includegraphics[width=0.24\textwidth,height=10cm,keepaspectratio]{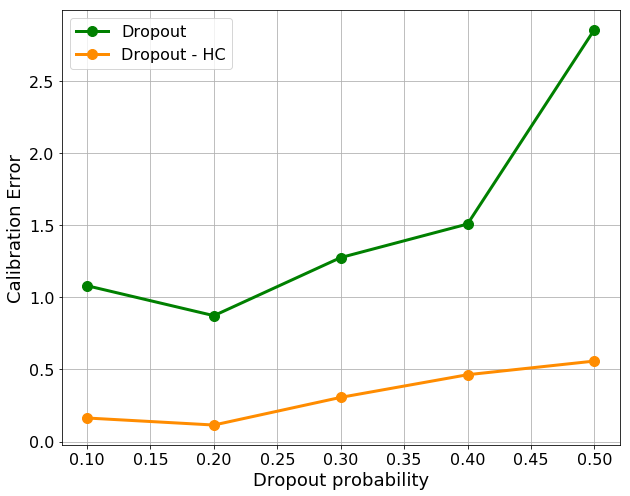}\label{fig:ce vs dp-boston}}
	\subfloat[S][Boston - Distribution of $\mathrm{\sigma}$]{\includegraphics[width=0.24\textwidth,height=10cm,keepaspectratio]{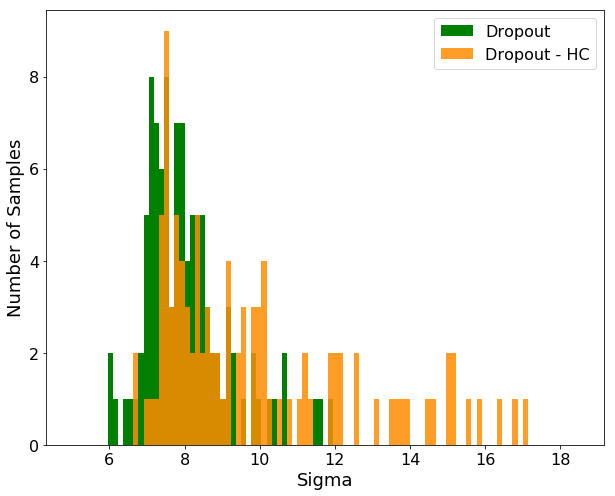} \label{fig:widths_boston}}
	\subfloat[S][Crime - Dropout Rate vs CE]{\includegraphics[width=0.24\textwidth,height=10cm,keepaspectratio]{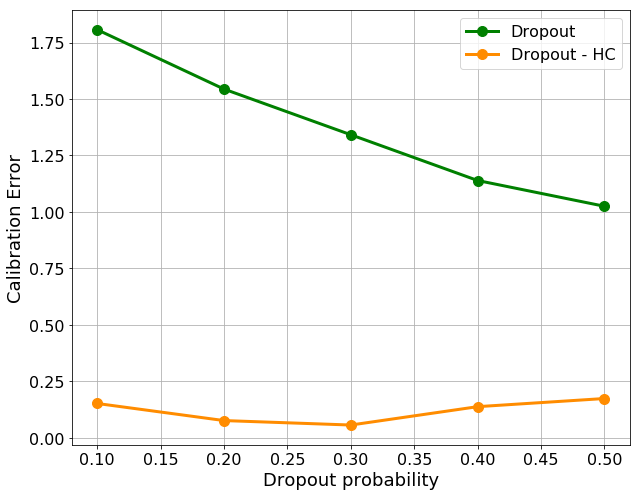}\label{fig:ce vs dp-crime}}
	\subfloat[S][Crime - Distribution of $\mathrm{\sigma}$]{\includegraphics[width=0.24\textwidth,height=10cm,keepaspectratio]{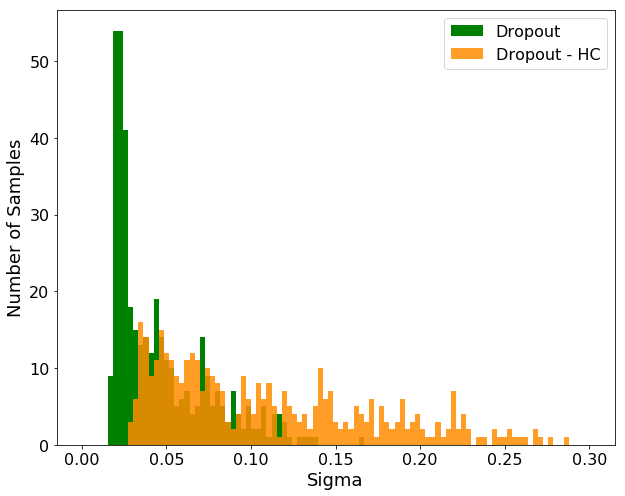} \label{fig:widths_crime}}
	
	\caption{Behavior of the proposed heteroscedastic calibration approach with the MC Dropout estimator on two datasets. }
	\label{fig:calib_performance}
\end{figure*}

\paragraph*{Deep Uncertainties:}While probabilistic Bayesian models are typically used for measuring uncertainties and reasoning about confidences, they are developed with simple modeling assumptions (e.g. Gaussian processes) and are known to be computationally expensive. Hence, a wide variety of approximate estimation strategies have been designed for neural networks. Examples include MC dropout~\cite{gal2016dropout}, concrete dropout~\cite{gal2017concrete}, deep ensembles~\cite{lakshminarayanan2017simple}, Bayes-by-backprop~\cite{ghahramani2015probabilistic} etc. 

In this paper, we consider two popular uncertainty estimation strategies to demonstrate the usefulness of heteroscedastic calibration: (a) \textit{MC Dropout}: This is a widely adopted technique for measuring epistemic uncertainties. In~\cite{gal2016dropout}, it was showed that deep networks with dropout applied before every weight layer are mathematically equivalent to a variational inference with deep Gaussian processes. This key result provides a Monte-Carlo style estimation technique that performs multiple forward passes for an input (with dropout) and produce mean/variance estimates from the multiple realizations, (b) \textit{Quantile-Based Estimator}: In~\cite{tagasovska2018frequentist}, the authors proposed a simultaneous quantile regression strategy to estimate aleatoric uncertainties. Let $F(\mathrm{y}) = P(\mathrm{y} \leq y)$ be the strictly monotonic cumulative distribution function of $\mathrm{y}$ assuming real values $y$ and $F^{-1}(\tau) = \inf \{y: F(\mathrm{y}=y) \geq \tau\}$ denotes the quantile distribution function for $0 \leq \tau \leq 1$. Quantile regression is aimed at inferring a desired quantile $\tau$ for the target $\mathrm{y}$, when the input $\mathrm{x} = x$, i.e., $F^{-1}(\tau | \mathrm{x} = x)$. This model can be estimated using the \textit{pinball} loss~\cite{koenker2001quantile}:
\begin{equation}
\mathrm{L}_{\tau}(\mathrm{y},\hat{\mathrm{y}}) = 
\begin{cases}
\tau(\mathrm{y} - \hat{\mathrm{y}}) ,& \text{if }\mathrm{y} - \hat{\mathrm{y}} \geq 0, \\
(1-\tau)(\hat{\mathrm{y}} - \mathrm{y}),              & \text{otherwise}.
\end{cases}
\end{equation}Recently, \cite{tagasovska2018frequentist} showed that the \textit{inter-quantile range} between appropriately chosen upper and lower quantiles can produce estimates of the aleatoric uncertainties. For a given $\alpha$,
\begin{align}
&\text{IQR}(\mathrm{x}; \tau^u, \tau^l) = \mathcal{F}(\mathrm{x}|\tau = \tau^u) - \mathcal{F}(\mathrm{x}|\tau = \tau^l),\label{eqn:iqr} \\
\nonumber&\tau^u = (1 + \alpha)/2, \tau^l = (1 - \alpha)/2.
\end{align}

%% file: approach.tex
We now develop the heteroscedastic calibration strategy while building predictive models. We argue that the same approach can be used to improve the calibration of any existing uncertainty estimator, thus making the resulting intervals meaningful for subsequent analysis. To this end, we utilize the heteroscedastic regression objective as a surrogate for calibration and entirely dispense the need for explicit recalibration as done in several recent approaches~\cite{kuleshov2018accurate,levi2019evaluating}. Furthermore, by relying on principled uncertainty estimation techniques, the intervals produced by our approach can be associated to specific uncertainty sources, unlike black-box methods such as HNNs~\cite{guo2017calibration}. Broadly, meaningful prediction intervals are expected to possess properties such as low entropy, good calibration, sharpness etc.~\cite{gneiting2007probabilistic}. However, it is not possible to incorporate these properties as constraints into the HNN training process. In contrast, our approach provides a principled way to enforce these properties through the use mathematically-grounded uncertainty estimators. As described in the previous section, we consider two different uncertainty estimation strategies to implement the proposed approach. Most existing techniques for uncertainty quantification in deep models operate on trained models to produce prediction intervals post-hoc. In contrast, our approach integrates the estimator into the training process and performs single-shot calibration through the heteroscedastic objective. We now describe the algorithms for augmenting two popular uncertainty estimators with hetereoscedastic calibration.

\subsection{Dropout-HC}
As showed in Algorithm \ref{algo-d}, following the MC dropout strategy, we perform multiple forward passes with dropout during every training iteration. More specifically, we repeat the forward pass $M$ times for each sample $\mathrm{x}_i$ to obtain predictions:
\begin{equation}
    \hat{\mathrm{y}}_i^j = \mathcal{F}(\mathrm{x}_i; \text{dropout} = p),
\end{equation}where $j$ is the index of the MC iteration. Subsequently, we obtain mean and variance estimates for the prediction using the $M$ realizations. Finally, we optimize the parameters $\Theta$ of the model $\mathcal{F}$ using the heteroscedastic objective in Eq. (\ref{eq:hnn}), based on the MC dropout estimates $(\mathrm{\mu}_i, \mathrm{\sigma}_i^2)$. Note that, by design, MC dropout measures the impact of model perturbations on the prediction. Consequently, upon optimization, the model parameters are updated such that the underlying estimator achieves improved calibration, while still taking into account only the model uncertainties.

\input{table.tex}

\subsection{Quantile-HC}
An interesting aspect of the proposed heteroscedastic calibration approach is it can be applied to any deep uncertainty estimator. In order to demonstrate that, we consider an aleatoric uncertainty estimator based on conditional quantiles~\cite{tagasovska2018frequentist}. From Algorithm \ref{algo-q}, it can be seen that the model is designed to output predictions at conditional quantiles $\tau^u$ and $\tau^l$, in addition to the mean estimate $\mathrm{\mu}$. We follow the strategy in Eq. (\ref{eqn:iqr}) to estimate the aleatoric uncertainty and obtain $\mathrm{\sigma}_i$ (for sample $\mathrm{x}_i$) for the heteroscedastic objective as:
\begin{align}
    &\mathrm{\sigma}_i = \frac{\hat{\mathrm{y}}_i^u - \hat{\mathrm{y}}_i^l}{2}; \hat{\mathrm{y}}_i^u = \mathcal{F}(\mathrm{x}_i|\tau = \tau^u), \hat{\mathrm{y}}_i^l = \mathcal{F}(\mathrm{x}_i|\tau = \tau^l).
\end{align}During optimization, we jointly obtain the parameters to estimate the mean as well as conditional quantiles, such that the intervals are well calibrated. Formally, we use the following loss function:
\begin{equation}
    \lambda_H\mathrm{L}_{\text{HNN}} + \lambda_u \mathrm{L}_{\tau^u} + \lambda_l \mathrm{L}_{\tau^l}
\end{equation}While the first term is similar to the Dropout-HC case, the last terms are included to estimate the conditional quantiles of the output distribution. The hyperparameter settings in our experiments were $\lambda_H = 0.75, \lambda_u = 1.0, \lambda_l = 1.0$.
\begin{figure}[!t]
	\centering
	\subfloat[S][Boston ]{\includegraphics[width=0.235\textwidth]{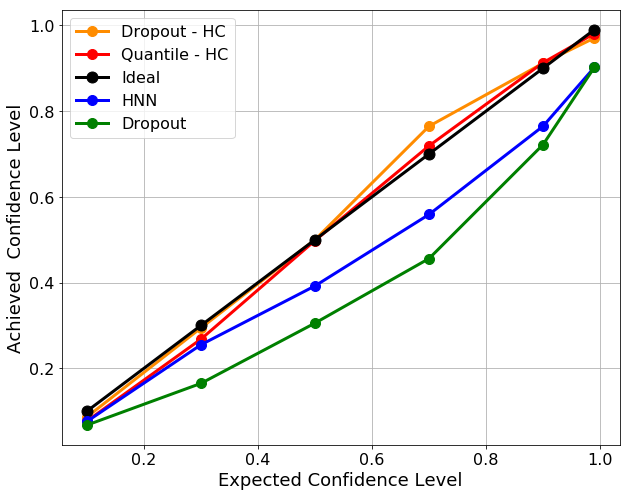}\label{fig:boston}}
	\subfloat[S][Crime]{\includegraphics[width=0.235\textwidth]{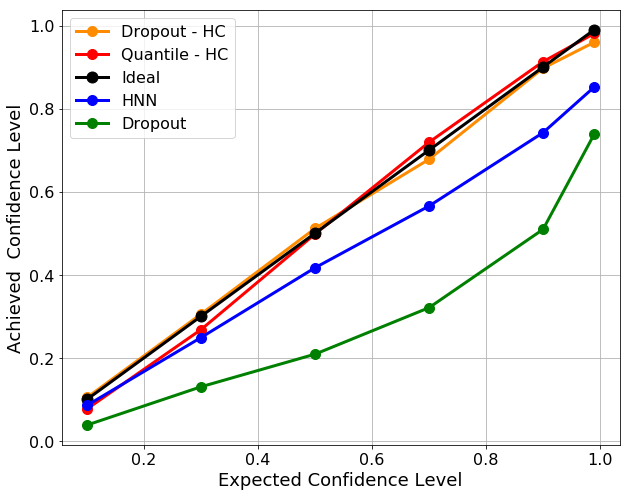} \label{fig: crime}}
	\caption{\textit{Expected vs Achieved Calibration Level} plots for the proposed Dropout-HC and Quantile-HC methods, in comparison to existing baselines.}
	\label{fig:calib}
\end{figure}

%% file: table.tex
\begin{table*}[] \label{table:logs}
\centering
\renewcommand*{\arraystretch}{1.3}
\begin{tabular}{|c|c|c|c|c|c|c|c|c|}
\hline
\multirow{2}{*}{\textbf{Dataset}} & \multicolumn{2}{c|}{\textbf{MC Dropout}} & \multicolumn{2}{c|}{\textbf{HNN}} & \multicolumn{2}{c|}{\textbf{Dropout-HC}}  & \multicolumn{2}{c|}{\textbf{Quantile-HC}} \\ \cline{2-9} 
                         & \textbf{CE}       & \textbf{RMSE}              & \textbf{CE}         & \textbf{RMSE}        & \textbf{CE}             & \textbf{RMSE}            & \textbf{CE}              & \textbf{RMSE}           \\ \hline \hline
Crime                    & 1.543    & 0.148             & 0.579      & 0.144       & \textbf{0.076} & 0.145           & 0.144           & \textbf{0.142} \\ \hline \hline 
Red Wine                & 1.955    & 0.636             & 0.185      & 0.653       & 0.151          & 0.619           & \textbf{0.096}  & \textbf{0.614} \\ \hline \hline 
White Wine              & 2.270    & \textbf{0.747}    & 1.092      & 0.768       & 0.269          & 0.758           & \textbf{0.158}  & 0.779          \\ \hline \hline 
Parkinsons               & 0.324    & 4.330              & 0.770            & {5.562}           & 0.836          & \textbf{4.322 }          & \textbf{0.173}           & 7.013          \\ \hline \hline 
Boston                   & 0.872    & 6.391             & 0.541      & 5.106       & \textbf{0.115} & 5.269           & 0.252           & \textbf{3.43}  \\ \hline \hline 
Autompg                 & 3.001    & 8.244             & 0.360      & 6.273       & 0.524          & 4.64            & \textbf{0.164}  & \textbf{2.826} \\ \hline \hline 
Superconductivity        & 0.495    & 11.971            & 0.912      & 10.820      & 0.194          & \textbf{10.092} & \textbf{0.068}  & 14.088         \\ \hline \hline 
Energy Appliance         & 0.213    & 87.275            & 2.315      & 86.692      & 0.209          & 89.564          & \textbf{0.159}           & \textbf{81.39}         \\ \hline 
\end{tabular}
\caption{Performance evaluation of the proposed approaches for the benchmark datasets. We report the RMSE of the mean predictions and calibration error of the prediction intervals.}
\end{table*}

%% file: results.tex
We now evaluate the proposed approach on several benchmark datasets, in terms of both generalization error and test calibration error. Note, though all our empirical studies are pertinent to regression, extending the proposed approach to classification tasks is straightforward. For all the experiments we used fully connected networks with five hidden layers with ReLU activation. For comparison, we show the results from two popular baselines, namely MC Dropout and HNN.  

\noindent \textbf{Setup}: We considered $8$ benchmark regression datasets from the UCI repository for our experiments. While the sample sizes of these datasets varied between $400$ and $\sim20,000$, the number of dimensions ranged between $7$ and $124$. In each of the cases, we used $80\%-20\%$ random data splits for training and evaluation. We use the RMSE metric to evaluate the quality of the mean estimator, and the CE metric from Eq. (\ref{eq:calib}), which quantifies the discrepancy between expected and achieved calibration levels, to measure the performance of the interval estimator. For both MC Dropout and Dropout-HC we used $p = 0.2$ and for Quantile-HC we set $\tau^u = 0.9, \tau^l = 0.1$.

\noindent \textbf{Results}: We find that the proposed approach produces highly accurate models that are also well calibrated, regardless of the uncertainty estimator used. As seen in Fig.~\ref{fig:calib}, both Dropout-HC and Quantile-HC approaches demonstrate superior calibration performace when compared to the existing baselines. In the \textit{Expected vs Achieved Calibraton Level} plots, perfect calibration corresponds to a diagonal line, and on both datasets showed, we observe the proposed approaches to be near-perfect. Next, we studied the behavior of the proposed approach with respect to uncertainty estimator design. In particular, we plot the impact of the dropout rate $p$ on the calibration error (CE) metric. Interestingly, though MC dropout is known to rely heavily on the choice of $p$, as demonstrated by large CE variability in Figures~\ref{fig:ce vs dp-boston} and~\ref{fig:ce vs dp-crime}, the proposed Dropout-HC is fairly robust with $p$. We also compared the distribution of $\mathrm{\sigma}$'s between MC Dropout and Droput-HC to find that, for the same value of $p$, the latter model produces a heavier tailed $\mathrm{\sigma}$ distribution. Finally, we report the RMSE and CE metrics for all datasets in Table 1. We find that the proposed approaches are significantly superior to the baselines in both accuracy and CE, thus evidencing that the heteroscedastic calibration leads to improved regularization.

%% file: conclusions.tex
We presented a strategy to enable uncertainty estimators to produce calibrated prediction intervals . More specifically, we showed that the heteroscedastic regression objective can be adapted to enable inherent calibration. We also developed the Dropout-HC and Quantile-HC algorithms for calibrating two popular uncertainty estimators from the literature. Though we used a simple Gaussian prior on the predictions, our empirical studies showed significant gains in both calibration and accuracy with benchmark regression tasks.